\RequirePackage[table]{xcolor}
\documentclass[journal,twoside,web]{ieeecolor}
\usepackage{jsen}
\usepackage{cite}
\usepackage[pdftex]{graphicx}
\usepackage{algorithmic}
\usepackage{graphicx}
\usepackage{textcomp}
\usepackage{wrapfig}
\usepackage{subcaption}
\usepackage{amsmath,amssymb}
\usepackage{algorithmic}
\usepackage[linesnumbered,ruled,vlined]{algorithm2e}
\usepackage{wrapfig}
\usepackage{graphics}

\newcommand{\frm}[1]{\langle #1\rangle}

\DeclareMathOperator*{\argmin}{argmin} % thin space, limits underneath in displays
\def\BibTeX{{\rm B\kern-.05em{\sc i\kern-.025em b}\kern-.08em
    T\kern-.1667em\lower.7ex\hbox{E}\kern-.125emX}}
\definecolor{abstractbg}{rgb}{0.89804,0.94510,0.83137}
\begin{document}

\title{On-line Optimal Ranging Sensor Deployment for Robotic Exploration}

\author{Luca Santoro, Davide Brunelli, \IEEEmembership{Senior Member, IEEE}, and Daniele Fontanelli, \IEEEmembership{Senior Member, IEEE}
\thanks{This work was supported by the Italian Ministry for Education, University and Research (MUR) under the program “Dipartimenti di Eccellenza (2018-2022)”. }
\thanks{Authors are with Department of Industrial Engineering, University of Trento, Trento, Italy (e-mail: \{luca.santoro, davide.brunelli, daniele.fontanelli\}@unitn.it).}
\thanks{ }
\thanks{This article has been accepted for publication in IEEE Sensors Journal. Citation information: L. Santoro, D. Brunelli and D. Fontanelli, "On-line Optimal Ranging Sensor Deployment for Robotic Exploration," in IEEE Sensors Journal, doi: 10.1109/JSEN.2021.3120889.}}

\IEEEtitleabstractindextext{%
\fcolorbox{abstractbg}{abstractbg}{%
\begin{minipage}{\textwidth}%
\begin{wrapfigure}[12]{r}{3in}%
\includegraphics[width=2.3in]{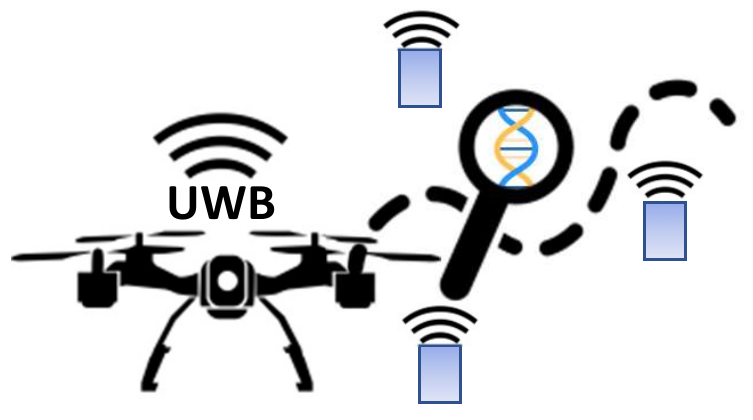}%
\end{wrapfigure}%
\begin{abstract}
Navigation in an unknown environment without any preexisting positioning infrastructure has always been hard for mobile robots. 
This paper presents a self-deployable ultra wideband UWB infrastructure by mobile agents, that permits a dynamic placement and runtime extension of UWB anchors infrastructure while the robot explores the new environment.
We provide a detailed analysis of the uncertainty of the positioning system while the UWB infrastructure grows. Moreover, we developed a genetic algorithm that minimizes the deployment of new anchors, saving energy and resources on the mobile robot and maximizing the time of the mission. 
 Although the presented approach is general for any class of mobile system, we run simulations and experiments with indoor drones.
Results demonstrate that maximum positioning uncertainty is always controlled under the user's threshold, using the  Geometric Dilution of Precision (GDoP).
\end{abstract}

\end{minipage}}}

\maketitle
\section{Introduction}

Mobile robotics, either terrestrial or aerial, have quickly registered incremental advances and interests from the industry and research community. Nowadays, they are pervasively applied in a variety of applications. Exploration of unknown environments has attracted an increasing attention due to their large application scenarios,  such as search and rescue missions~\cite{Rahman2018}, disaster recovery~\cite{SeminaraF17mesas}, planetary exploration~\cite{Onge2020}, photogrammetry~\cite{Dawei2020}, aerial inspection and monitoring of buildings and structures~\cite{He2019}~\cite{Bacco2020}, agriculture~\cite{Murugan2017} and predictive maintenance~\cite{William2018}. 
Localisation and positioning capabilities are primary features for any autonomous exploration system. According to the application scenarios, several solutions can be  used. The capability of positioning in an absolute reference system, usually with the GPS signal~\cite{Wang2020}, is one of the most used techniques. 
However, many robot exploration activities are in GNSS-denied environments, such as indoor. In such challenging cases, alternative positioning methods are usually considered, e.g., visual-SLAM~\cite{Carrio2020}~\cite{Silva2015}, laser scanners~\cite{Lin2011}.
Some of these techniques require non-negligible computing resources, work preferably in information-rich environment, and cannot guarantee a maximum target uncertainty (e.g., SLAM)~\cite{Rodri2018}. Others, instead, have limited computational burden and can compute positioning under controlled uncertainty. Nevertheless, this class of solutions usually requires instrumented infrastructure in the surrounding with active or passive markers. Examples of this category are Radio Frequency (RF) active beacons for Radio Signal Strength Identification (RSSI)~\cite{Yang2020} or Ultra Wide Band (UWB)~\cite{MagnagoCPPF19iros}~\cite{Li2020}~\cite{Santoro2021}, while for passive solutions we can mention visual markers~\cite{NazemzadehFMP17tmech} or passive RFID tags~\cite{MagnagoPBTMNMF20tim}. The nature of the technology and the sensors embedded into the environment determine uncertainty during the exploration. % (e.g., RSSI techniques are, in general, far less precise than UWB solutions).

Providing positioning measurements with limited uncertainty for autonomous robot navigation is hard if an absolute reference as the GPS is not available. The achievable positioning performance depends both on the specific technology used by the sensing devices and on the algorithm defined for the placement of such devices~\cite{MagnagoPPFM19tim}.  When RF ranging sensors are considered, two different approaches are usually implemented to achieve the optimal placement in an unknown environment:
\begin{enumerate}
    \item \textbf{Off-line}. The environment is analyzed, e.g., using statistics about the navigation paths, and the placement positions are determined to guarantee the desired target uncertainty. For example, in~\cite{Bulusu2001},  three off-line algorithms are assessed and compared to find the candidate points of an additional beacon that maximizes the accuracy of the localisation service over the entire region. 
    \item \textbf{Online}. In this case, the environment may not be known upfront, and the anchors are deployed on-demand, e.g., when the localisation uncertainty approaches the maximum tolerable value. For instance, in~\cite{Gurkan2014}, the robots use two different strategies to place a new sensor in the environment: measure the average of RSSI, and place the new sensor when this value falls under a predetermined value or based on a fixed distance.
\end{enumerate}

Recently, ultra wideband UWB signals technology has been quickly confirmed as an effective and cheap solution for positioning problems. 
Different works are developed using UWB to decrease the target uncertainty. For example, in~\cite{Yoon2017}~\cite{Yang2021}~\cite{Vandermeeren2021}, Kalman filters are used to fuse UWB and Inertial Measurement Units (IMU) for improving the position estimation and mitigating the problem of Non-Line-of-Sight (NLoS).  In~\cite{Wang2021}, the poor estimation along the vertical axis, which is a known weakness of the UWB infrastructures, is mitigated using inter-vehicle distance. In~\cite{Kolakowski2020}, a combination of LiDAR (Light Detection and Ranging) and UWB is investigated, exploiting LiDAR information to improve the UWB results. 

The main characteristic of UWB technology is to use message exchange between mobile and fixed nodes. Typically, the mobile nodes are mounted on the robot's chassis, while the fixed nodes, named anchors, build up an infrastructure with known geometric characteristics. 
The fixed structure of nodes is usually deployed before starting any operations in the environment~\cite{Shi2019}, hence adopting an off-line placement procedure. Although it seems simple, off-line placement is time-consuming and critical because any fault in this phase or anchor position uncertainty seriously influences the positioning system's precision.

To overcome infrastructure setup inaccuracies and provide a positioning system also for unstructured environments, where off-line analysis is inapplicable,  we developed a method for dynamic placement and runtime extension of the infrastructure anchors. 
In our work, while exploring the environment, the mobile robot deploys new anchors to strengthen the infrastructures. Thus the ranging sensors are self-deployable and will extend the positioning reference at runtime during the robot exploration.  Notice that this marks a striking difference with respect to the known literature. Indeed, existing solutions, e.g., \cite{ans121, ans122}, cannot change the nodes infrastructure at runtime based on robot needs nor can adequately leverage the ratio of information versus uncertainty that a new added anchor injects in the multilateration problem. Moreover, our solution is robot-centered: existing solutions usually try to optimize the entire region as a whole, with evident computational burden issues and difficulties in unknown or partially known environments, while our solution is extremely light in terms of computing power and can be computed onboard the vehicle while it explores the (possibly unknown) environments. In particular, our solution proposes an online-incremental algorithm based on a genetic approach to solve the constrained optimization problem, which finds the most convenient placement for new anchors and reduces the number of deployments. The algorithm keeps the maximum target uncertainty below the user requirement, which is based on the Geometric Dilution of Precision (GDoP). It has to be noted that the proposed solution works with any metric able to express the positioning uncertainty, but the GDoP comes handy for this purpose~\cite{FengSLD15sensors}.
 
In this work, we focus on positioning accuracy problems and not strictly on robot localisation, which requires the analysis of the problem's observability and the model of the robot dynamics~\cite{Fontanelli21imm}. We present how to use the proposed approach for a generic class of robot dynamics (e.g., ground or aerial vehicles), mainly focusing on positioning uncertainty.

The paper is organised as follows. Section~\ref{sec:Background} describes the placement metric adopted and the formulation of the problem. Section~\ref{sec:AnchorDeployment} presents the solution based on a genetic algorithm to optimize the chosen metric and the optimal deployment manoeuvres, while Section~\ref{sec:UncertaintyAnalysis} investigates the position uncertainty of the proposed deployment algorithm. Section~\ref{sec:SimulationResults} discuss the simulation and experimental results. Finally, Section~\ref{sec:Conclusion} concludes the article and proposes future research directions.

\section{Background}
\label{sec:Background}
\begin{figure}[t]
    \centering
    \begin{tabular}{c}
        \includegraphics[width=1\columnwidth]{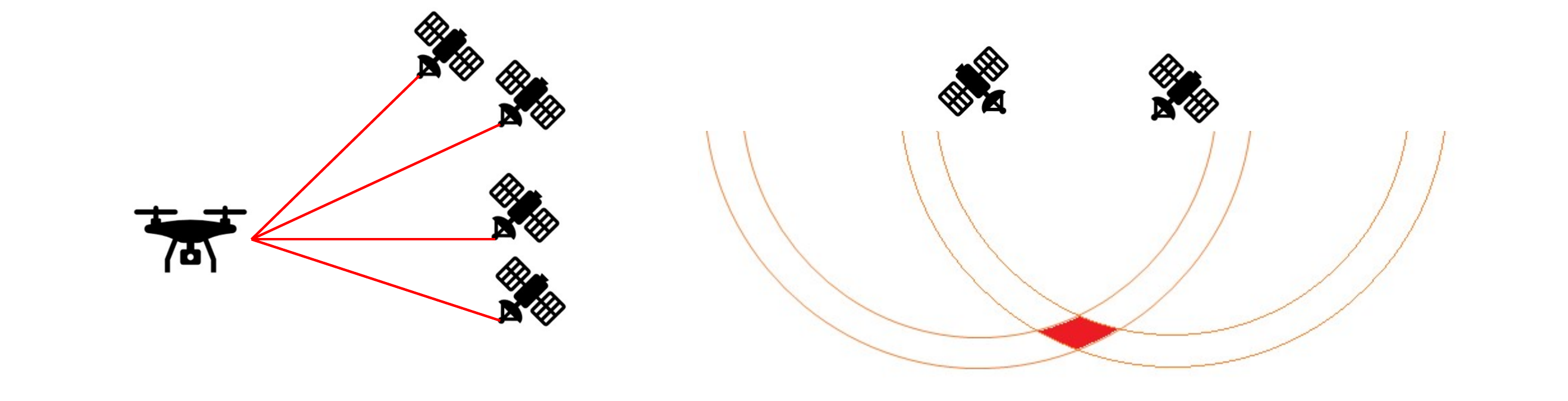} \\
        (a) \\
        \includegraphics[width=1\columnwidth]{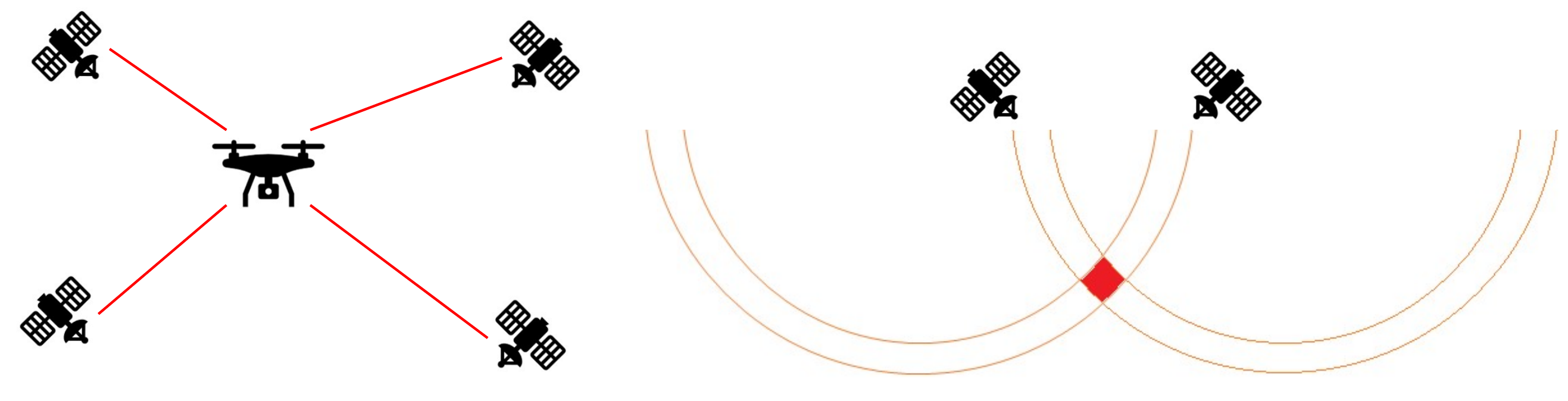} \\
        (b)
    \end{tabular}
    \caption{Circular sector intersections from (a) bad or (b) good distributions of satellites}
    \label{fig:GDoPDistributions}
\end{figure}

GDoP is a metric adopted to quantify the precision and accuracy of the data received from GPS satellites, which is now being adopted to the wider set of generic positioning system~\cite{YangYong,Sharp}. This metric indicates how well the satellites are geometrically organized. The lower the value, the better is the position accuracy~\cite{Wu}. A graphical representation of a poor or good geometric configuration is given in Figure~\ref{fig:GDoPDistributions}.
%-%
%-%
GDoP is proportional to the ratio between the range error and position error~\cite{Dana1997}, thus it is inversely proportional to the volume formed by the vectors from user to satellites and the number of satellites.
Given the distance $\rho_i$ from the $i$-th anchor and assuming that all the ranging measurements have the same finite variance (hence, the homoscedastic property is satisfied), we define the variance associated to the ranging as $\sigma_{\rho}^2$ and, from~\cite{MILLIKEN1978}, the covariance matrix of the positioning error is
\begin{equation}
\label{eq:CovGDoP}
C= \sigma_{\rho}^2
\begin{bmatrix}
\sigma_{xx}^2 & \sigma_{xy}^2 & \sigma_{xz}^2 & \sigma_{xt}^2\\
\sigma_{yx}^2 & \sigma_{yy}^2 & \sigma_{yz}^2 & \sigma_{yt}^2\\
\sigma_{zx}^2 & \sigma_{zy}^2 & \sigma_{zz}^2 & \sigma_{zt}^2\\
\sigma_{tx}^2 & \sigma_{ty}^2 & \sigma_{tz}^2 & \sigma_{tt}^2\\
\end{bmatrix} ,
\end{equation}
where  $\sigma_{\rho}^2 \sigma_{xx}^2, \sigma_{\rho}^2 \sigma_{yy}^2, \sigma_{\rho}^2 \sigma_{zz}^2$ represent the variance of the estimated location along the corresponding axes and $\sigma_{tt}^2$ is the time offset of the receiver. Sub-metrics can be defined from~\eqref{eq:CovGDoP} by adopting the trace on different sub-matrices, such as
\begin{equation}
\label{eq:DifferentMetrics}
  \begin{cases}
    \mbox{HDoP}=\sqrt{\sigma_{xx}^2+\sigma_{yy}^2} ,\\
    \mbox{VDoP}=\sigma_{zz}^2 , \\
    \mbox{PDoP}=\sqrt{\sigma_{xx}^2+\sigma_{yy}^2+\sigma_{zz}^2} , \\
    \mbox{GDoP}=\sqrt{\sigma_{xx}^2+\sigma_{yy}^2+\sigma_{zz}^2+\sigma_{tt}^2} , \\
  \end{cases}
\end{equation}
where HDoP, VDoP and PDoP are the Horizontal, Vertical and Position Dilution of Precision, respectively, all derived from the GDoP.
We moved these metrics to UWB infrastructures. Thus the position estimate of a receiver (called {\em tag}) in a generic three-dimensional space requires at least four UWB devices (called anchors). In contrast to~\eqref{eq:CovGDoP}, the time $t$ is not of interest for UWB ranging system because the propagation time of the signal is directly used for the time-of-flight measurement~\cite{LazzariBNL17sensors}, hence the last column of $C$ will be neglected. Consequently, the PDoP metric in~\eqref{eq:DifferentMetrics} is used in place of the GDoP. 

\subsection{Problem formulation}
\begin{figure}[t]
    \centering
    \includegraphics[width=0.7\columnwidth]{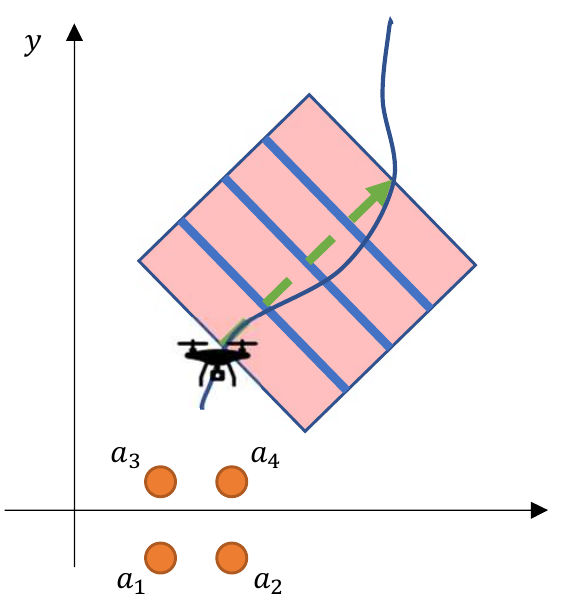}
    \caption{A graphical representation of the problem statement. Example of subareas $S(i+(j-1)r/n,jr/n)$,  with j=1,...,n, using the described simplified approach.}
    \label{fig:Problem_formulation}
\end{figure}

The contribution of this paper is to derive an optimal self-deployment on-line solution for UWB anchors during exploration problems and using the PDoP metric to define the target uncertainty. The optimal reduction of positioning uncertainties is tailored to the robotic platforms' requirements,  saving onboard hardware and computation resources and time. The algorithm calculates the minimum number of anchors to deploy during the mission to accomplish the robot goal.  More formally, let us consider the situation depicted in Figure~\ref{fig:Problem_formulation}.
%-%
%-%
We assume that the anchors can only be deployed on the $X_w\times Y_w$ plane of the right-handed reference frame $\frm{W} = \{X_w, Y_w, Z_w\}$, since the altitude of the placement is assumed to be not controllable. Notice that assuming no knowledge about the environment, we consider the worst possible conditions for the $z$ coordinates, i.e., that the multilateration algorithm is applied using coplanar anchors (flat terrain), hence we are assuming very poor VDoP. We denote with $a_i = [X_i,Y_i]^T$ the known coordinates of the anchor in $\frm{W}$ and projected on the plane $X_w\times Y_w$. Moreover, given $\mathcal{A}_k$ the set of all the anchors $a_i$, we define with $\overline{\mathcal{A}}_{k,n}$ as the set of all the combinations of $n$ anchors in $\mathcal{A}_k$.  Therefore, $\mathcal{D}_k(s)\in\overline{\mathcal{A}}_{k,4}$ denote the set of $4$ anchors attaining the minimum value of PDoP in a certain position $s$. As a consequence, given:
\begin{itemize}
    \item A sampling time $T_s$, which is induced by the sampling time of the available anchors;
    \item A planned exploration path $S_p=\{q_i\}_{i=1}^h$ is a set of $h$ viapoints on the plane, i.e. $q_i=[x_{q_i},y_{q_i}]^T$;
    \item The actual position of the robot $s_k=[x_k,y_k]^T$ at time $k T_s$, supposed to be projected on the plane $X \times Y$, while $S_k = \{s_i\}_{i=0}^k$;
    \item An initial set $\mathcal{A}_0$ of $4$ anchors that are in communication with the robot;
    \item A set of the overall deployed anchors $\mathcal{A}_k$ up to time $k T_s$;
    \item A maximum tolerable value ${p}^m$ of the PDoP along the exploration path;
    \item A maximum distance ${\rho}^m$ from an anchor to retrieve the ranging measurement;
    \item A PDoP function $g(\mathcal{D}_k(s_k), s_k)$ computed on the position $s_k$ given the anchors $a_i\in\mathcal{D}_k$;
\end{itemize}
the goal is to guarantee the existence of at least four anchors $\mathcal{D}_k\subset\mathcal{A}_k$ at time $k T_s$, such that the UWB positioning system can provide a PDoP $g(\mathcal{D}_k(s_k), s_k) \leq {p}^m$, $\forall s_k$ during the exploration while using the ranging data $\rho_{i,k} = \|s_k - a_{i}\| \leq {\rho}^m$. 
To this extend, we define two problems:\\
i) the first is the {\em Optimal placement problem} (OPP)
\[
\begin{aligned}
    & \min \#\mathcal{A}_k \mbox{ s.t. } \\
    & \exists \mathcal{D}_k(q_i)\subseteq\mathcal{A}_k \mbox{ with } g(\mathcal{D}_k(q_i), q_i) < {p}^m, \forall q_i \in S_p ,
\end{aligned}
\]
where of course $\#\mathcal{A}_k$ are the number of elements in $\mathcal{A}_k$. \\
ii) The second, named {\em Optimal Exploration and Placement Problem} (OEPP), is based on OPP and defined on the actual robot positions $s_k$,  instead of the planned positions $q_i$. The difference between the two problems is that OPP refers to the nominal robot trajectory, while OEPP considers all the maneuvers needed to deploy the new anchors.

In this paper, we will make explicit reference to a particular class of robots, namely Unmanned Aerial Vehicles (UAVs), even though the solution remains of general validity. It is worthwhile to mention that the algorithm is totally agnostic about the planner used to synthesize the robot path. One of the most used exploration methods is the sampling-based algorithm such as RRT~\cite{karaman2011sampling}. Moreover, notice that the robot should entirely cover the exploration path at least once, i.e. $\exists k\in\mathbb{N}$ such that $s_k = q_i$, $\forall q_i\in S_p$.

\section{Anchors Deployment Algorithm}
\label{sec:AnchorDeployment}
At a first glance, OPP may appear a trivial problem that could be solved by computing $g(\mathcal{D}_k(s_k), s_k)$ at time $k T_s$ for the positions $s_k = q_i$ and place a new set of $4$ anchors on the same pattern of Figure~\ref{fig:Problem_formulation}, when either $g(\mathcal{D}_k(s_k), s_k) = {p}^m$ or $\rho_{i,k} = {\rho}^m$ for some $a_i\in\mathcal{A}_k$. Then the robot starts over. 
However, we observe three different problems with this approach (which is inspired by~\cite{BEINHOFER20131060} applied to ground wheeled vehicles): 
\begin{enumerate}
    \item While placing $4$ anchors satisfies the sufficient requirement for positioning, it does not guarantee that it is the only possible deployment. %given for granted that this is a necessary condition as well;
    \item When the UAV places the $4$ anchors, the PDoP should be kept under control on the placement trajectory as well, hence a PDoP threshold $p^\ast < {p}^m$ should be considered; 
    \item The next location to place the anchors may be a function of the future exploration path viapoints $S_p$, thus making some locations more favorable than others.
\end{enumerate}
To address these issues, we propose the Genetic Anchor Node Placement (GANP) algorithm, which comprises a prediction of the PDoP function along the future path positions with a finite horizon $r$ and then compute the most favourable locations using a Genetic Algorithm (GA).  More precisely, let us consider the robot is in position $s_k = q_i$, the algorithm starts by evaluating if $\exists q_j\in\{q_i,q_{i+1},\dots,q_{i+r}\} = S_{p}^{(i,r)}\subset S_p$ such that $g(\mathcal{D}_k(q_j), q_j) > {p}^m$. In such a case, we need to add at least one anchor. We can then grow a region $\mathcal{S}_{p}^{(i,r)}$ around the path portion $S_{p}^{(i,r)}$ of width $w$ by simply taking the local perpendicular to the path of length $w$ passing through each position $q_j = S_{p}^{(i,r)}$. However, 
to avoid a placement that concentrates the anchors in nearby positions, we split the finite horizon $r$ in $n$ subsets of $r/n$ points and then we could define non-overlapping regions $\mathcal{S}_{p}^{(i,r/n)},\dots,\mathcal{S}_{p}^{(i+r-r/n,i+r)}$ each hosting at most one new anchor. To simplify the subareas splitting, we simply take the line joining $q_i$ and $q_{i+r}$ and consider it as an approximation of the path, thus simplifying the searching regions comprised in $\mathcal{S}_{p}^{(i,r)}$ as sketch in Figure~\ref{fig:Problem_formulation}.
Of course, the width $w$ of the searching region, the forecasting horizon $r$, and the number of sub-paths $n$ plays a crucial role in the algorithm performance, hence a tuning procedure is presented in Section~\ref{sec:SimulationResults}.

\begin{figure}[t]
    \centering
    \includegraphics[width=0.7\columnwidth]{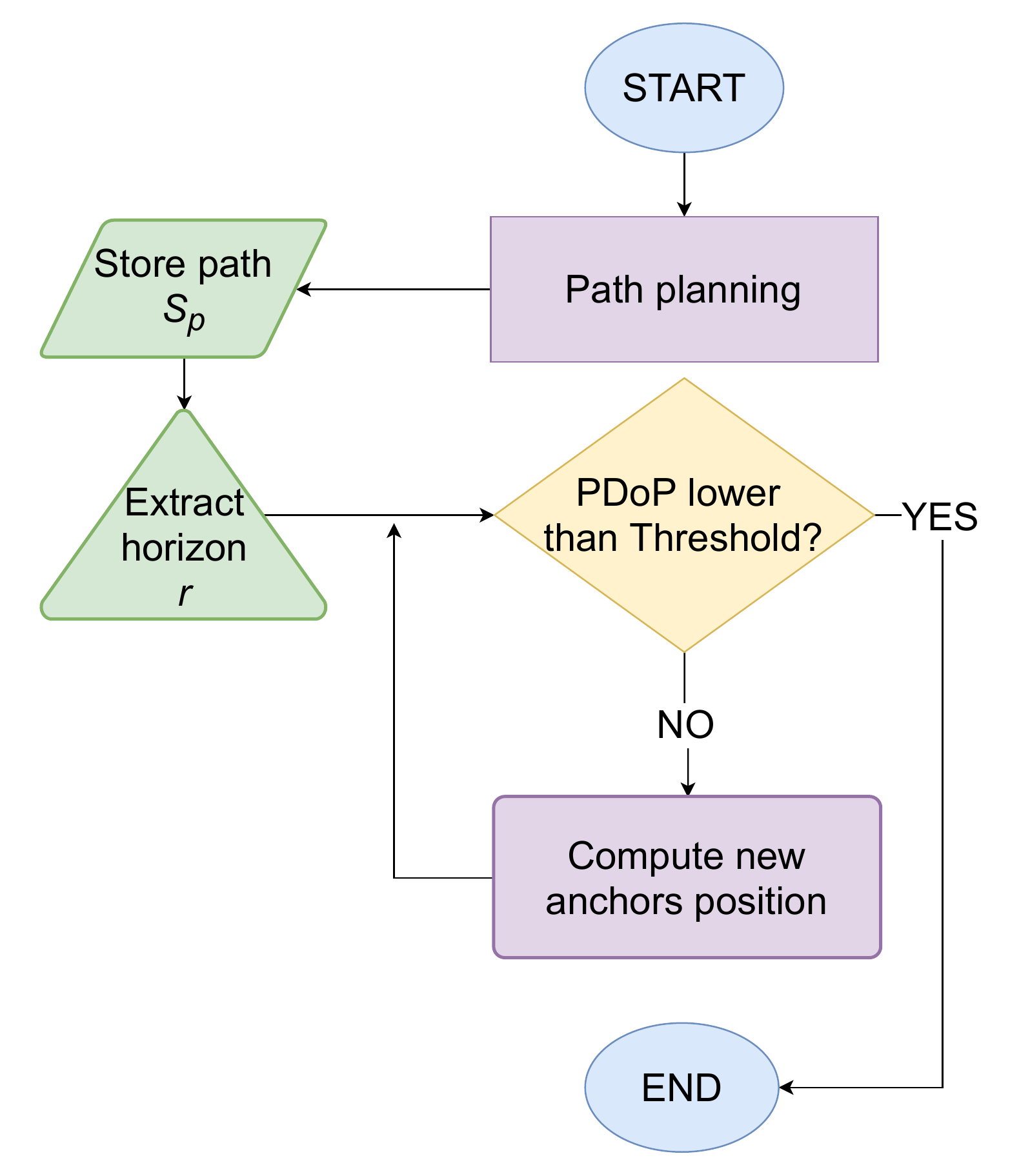}
    \caption{Flowchart of GANP algorithm.}
    \label{fig:Overview of GANP algorithm}
\end{figure}

A flowchart of the GANP algorithm is depicted in Figure~\ref{fig:Overview of GANP algorithm}.
%-%
%-%
Whenever one or more anchors should be placed according to the PDoP function, a list of the possible deployment coordinates is computed. The main idea is to allow at most one anchor in each subarea $\mathcal{S}_{p}^{(i+(j-1)r/n,j r/n)}$, with $j = 1,\dots,n$.
Suppose four anchors result placed in the $j$-th area to achieve an optimal PDoP value. In that case, it means that the considered $j$-th area is too detached (i.e., far)  from the infrastructure, then $j-1$-th area is considered forcing the algorithm to generates at least one additional anchor.
 
The GA {\em fitness function} considers the optimal PDoP values for each point $q_j\in S_p^{i,r}$ by determining the set $\mathcal{D}_k(q_j)$. The PDoP quantities are then stored into a list and weighted according to the distance from $q_i$: the more $\|q_j - q_i\|$ is larger, the higher is the weight. These values are then summed up and constitute the objective function to minimize, i.e.
\[
    Y = \sum_{i=1}^{r}g(\mathcal{D}_k(s_k), s_k)~log(\|q_j - q_i\|) .
\]
Notice that the weighting mechanism pushes the new possible anchors deeper along the exploration path, maximizing the effect of the coverage and ensuring the minimum number of anchors for the considered subset $S_p^{i,r}$. The GA {\em constraint function} checks the following three conditions for each generated possible anchor location:
\begin{itemize}
    \item The position of the generated anchor should have a PDoP value below the maximum threshold $p^m$;
    \item The path joining the robot position and the anchor candidate location should have a PDoP below $p^m$ as well;
    \item The PDoP of all the positions in $S_p^{i,r}$ should be below $p^m$.
\end{itemize}
Notice that the GANP algorithm ensures the optimality of the OPP only. To extend the results to the OEPP, the nature of the deploying maneuver should be taken into account. 

\subsection{Deploying Manoeuvres}

\begin{figure}[t]
    \centering
    \includegraphics[width=0.8\columnwidth]{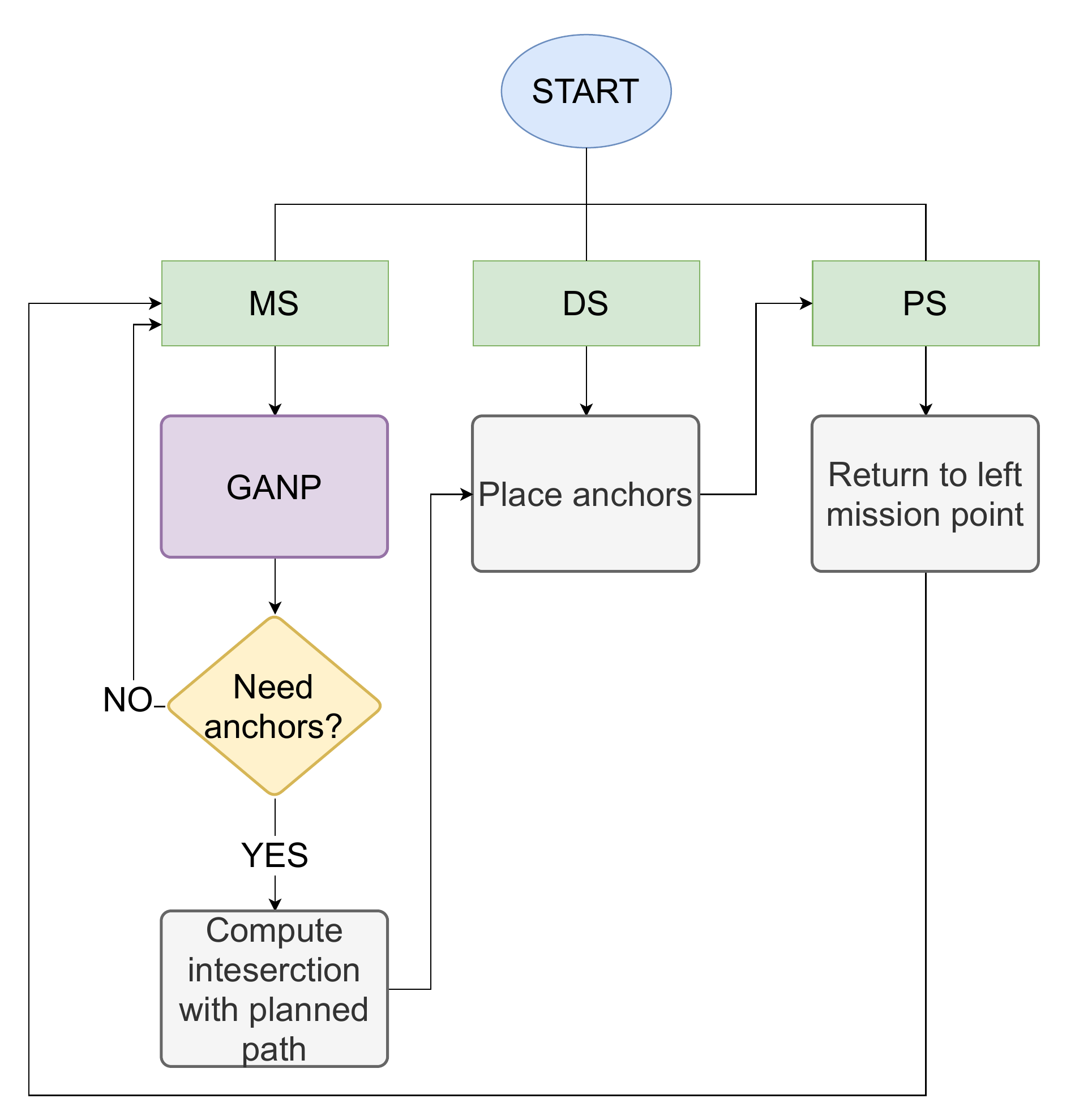}
    \caption{Overall logic that govern the behaviour of the drone during its mission. }
    \label{fig:Overall logic}
\end{figure}

\begin{figure}[t]
     \begin{subfigure}[b]{0.48\linewidth}
        \centering
        \includegraphics[width=0.8\linewidth]{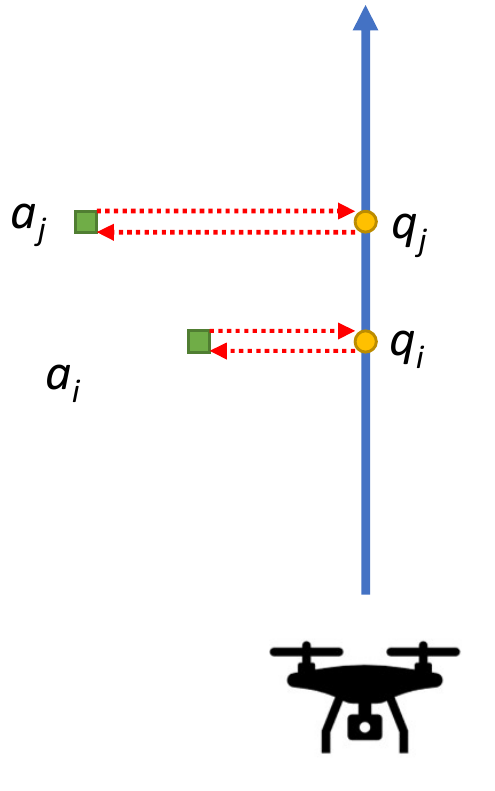}
        \caption{Deployment path generation, each anchor has its relative intersections point.}
        \label{fig:Placing single}
     \end{subfigure}
     \hfill
     \begin{subfigure}[b]{0.48\linewidth}
        \centering
        \includegraphics[width=0.73\linewidth]{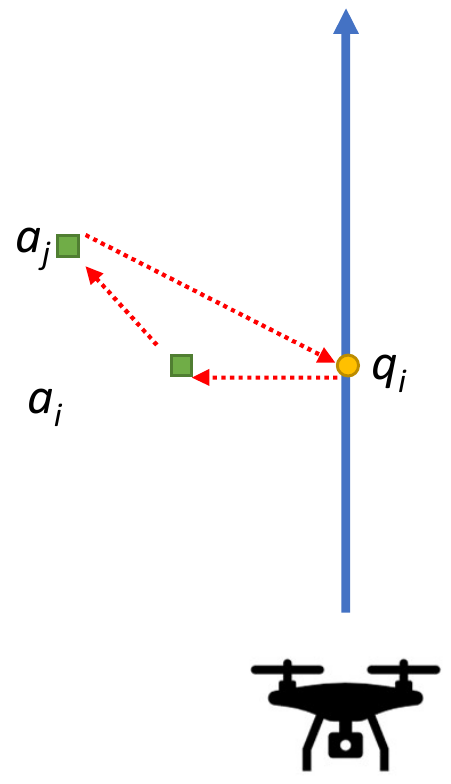}
        \caption{Deployment path generation, from first intersection point, all anchors in list are deployed.}
        \label{fig:Placing sequential}
     \end{subfigure}
      \caption{Deployment manoeuvres followed in DS and PS.}
\end{figure}

The GANP algorithm ensures that the value of PDoP never exceeds the maximum target value ${p}^m$. In fact, the algorithm governs the UAV controlled behaviour based on three states of a Finite State Machine. The UAV starts in {\em Mission State} (MS) where it follows the exploration path. When the condition is violated on the path horizon $r$, the optimal position of the anchor is determined, and the UAV switches to the {\em Deployment State} (DS). The UAV stores the last point reached along the mission path, say $q_i\in S_p$, and follows the shortest path towards the deploying location. After the anchor is positioned, the UAV either continues on the placement (if convenient, as described in the rest of this section) or it switches to the {\em Placed State} (PS). Here, a return-path to $q_i$ is generated and followed. When the robot reaches $q_i$ either switches back to DS (if additional anchors should be placed, or returns to MS, where the exploration continues. This motion pattern is pursued until the last point of the mission is reach, where the UAV decides which action to perform:
\begin{itemize}
    \item Landing (or stopping) and becoming an integral part of UWB positioning infrastructure with its tag that switches to an anchor. This action can be fired by the battery level when it falls below a certain threshold;
    \item Continue the exploration mission, selecting a new exploration area with a new synthesized  path and  executing the described process;
    \item Alternatively, the robot can move back to the starting position, increasing the accuracy of the placed anchors.
\end{itemize}
The flowchart of the depicted algorithm is reported in Figure~\ref{fig:Overall logic}.
%-%
%-%
The path followed in the DS and PS states is crucial for any vehicle autonomy, especially when UAVs are considered. Therefore, the maneuver should take the shortest. While the placement path for a single anchor is straightforward, i.e., it is sufficient to move along the local perpendicular segment with respect to the planned path (see Figure~\ref{fig:Placing single}), the placement of more than one anchor may be tricky.  
%-%
%-%
When the new anchor locations $a_i$ and $a_j$ are determined, two possible strategies are considered. The first is reported in Figure~\ref{fig:Placing single}: the back-and-forth motion is adopted whenever the UAV reaches an intersections points on the path (first $q_i$, then $q_j$). The second more involved situation is depicted in Figure~\ref{fig:Placing sequential}, where the UAV starts the deployment manoeuvres from the first intersection point, $q_i$, and sequentially place all the anchors before going back to the point $q_i$. The selection of the two strategies is made on the fly by comparing the perimeter of standard geometric shapes (recall that at most $4$ anchors should be placed at once), with the constraint that the manoeuvre in DS starts and ends in the same point $q_i$ (to cover the entire exploration path). For example, the path followed in DS for Figure~\ref{fig:Placing sequential} is shorter than the path of Figure~\ref{fig:Placing single}, thus it will be selected.
It is now evident that, by embedding this manoeuvre generation in the {\em constraint function} of the GA, the OEPP problem is solved requiring the limit of the PDoP to be satisfied along the shortest deployment manoeuvres.
\section{Uncertainty analysis}
\label{sec:UncertaintyAnalysis}

In this section, we first present the explicit derivation of the PDoP function $g(\mathcal{D}_k(s_k), s_k)$ and an analysis of the positioning uncertainty accounting for the incorrect anchor deployment.

\subsection{Position Dilution of Precision}

As described in Section~\ref{sec:Background}, the PDoP function used in this paper $g(\mathcal{D}_k(s_k), s_k)$ is a function of the anchor locations $\mathcal{D}_k(s_k) = \{a_{i_1}, \dots, a_{i_m}\}$ and of the point $s_k$ considered. In particular, defining with
\begin{equation}
\label{eq:PDoPFun}
P = \begin{bmatrix} 
        \frac{x_k - X_{i_1}}{\rho_{i_1,k}} & \frac{y_k - Y_{i_1}}{\rho_{i_1,k}}\\ 
        \vdots & \vdots\\
        \frac{x_k - X_{i_m}}{\rho_{i_m,k}} & \frac{y_k - Y_{i_m}}{\rho_{i_m,k}}
    \end{bmatrix} ,
\end{equation}
the Jacobian of the ranging function~\eqref{eq:rangingnoise} and denoting with $Q$ the covariance matrix of the positioning error
\[
    Q = \sigma_{\rho}^2 (P^T P)^{-1} = \sigma_{\rho}^2 \begin{bmatrix} 
            {\sigma_{xx}}^2 & {\sigma_{xy}}^2\\ 
            {\sigma_{yx}}^2 & {\sigma_{yy}}^2
    \end{bmatrix} ,
\]
the PDoP function $g(\mathcal{D}_k(s_k), s_k) = \sqrt{{\sigma_{xx}}^2 + {\sigma_{yy}}^2}$.

\subsection{Anchor deployment uncertainty}
\label{subsec:AnchorDeployment}

The position of the UAV is computed using multilateration on distance measurements. The ranging measurements are collected by means of an UWB infrastructure, using a Single Side Two-Way-Ranging (SS-TWR) communication protocol. Assuming $n$ UWB anchors in known positions $a_i = [X_i, Y_i]^T$, $i = 1,\dots,n$, the ranging measurement from the $i$-th anchor at time $k T_s$ is defined as
\begin{equation}
\label{eq:rangingnoise}
    \Bar{\rho}_{i,k} = \rho_{i,k} + \epsilon_{i,k} = \sqrt{(x_k - X_i)^2 + (y_k - Y_i)^2} + \epsilon_{i,k} ,
\end{equation}
where $\epsilon_{i,k}$ is the ranging measurement uncertainty, usually considered as a white sequence with zero mean and variance $\sigma_\rho^2$ for all the anchors. Computing the difference of the squares of the distances $\Delta_{ij,k} = \Bar{\rho}_{i,k}^2 - \Bar{\rho}_{j,k}^2$ from at least three anchors and using the same solution reported in~\cite{Fontanelli2021}, it is possible to derive the robot position estimates using a Weighted Least Squares (WLS) solution as
\begin{equation}
\label{eq:CovGDoP0}
        \hat s_k = \begin{bmatrix} \hat{x}\\ \hat{y} \end{bmatrix} = \frac{1}{2}(A^{(n)^T}N_k^{(n)^{-1}}A^{(n)})^{-1}A^{(n)^T}N_k^{(n)^{-1}}h_k^{(n)} ,
\end{equation}
where $h_k^{(n)}$ is the vector of the indirect measurements $\Delta_{ij,k}$ and anchor positions, $A^{(n)}$ is a matrix containing the known anchor positions, while
\begin{equation}
\label{eq:Nk}
    N_k^{(n)} = \sigma_{\rho}^2 
    \begin{bmatrix}
    \rho_{1,k}^2 + \rho_{2,k}^2 & \rho_{1,k}^2 & \ldots & \rho_{1,k}^2 \\
    \rho_{1,k}^2 & \rho_{1,k}^2 + \rho_{3,k}^2 & \ldots & \rho_{1,k}^2 \\
    \vdots  & \vdots  & \ddots & \vdots \\
    \rho_{1,k}^2 & \rho_{1,k}^2 & \ldots & \rho_{1,k}^2 + \rho_{n,k}^2
    \end{bmatrix} ,
\end{equation}
the covariance matrix of the measurements, which is a function of the actual distances $\rho_{i,k}$. The robot position uncertainty $\tilde s_k = \hat s_k - s_k$ derived from~\eqref{eq:CovGDoP0} has, hence, the following multilateration covariance matrix
\begin{equation}
    \label{eq:CovGDoP1}
    \Xi^{(n)} = (A^{(n)^T}N_k^{(n)^{-1}}A^{(n)})^{-1} =
    \begin{bmatrix}
    \sigma_{x,k}^2 & \sigma_{xy,k}^2 \\
    \sigma_{yx,k}^2 & \sigma_{yy,k}^2 \\
    \end{bmatrix} ,
\end{equation}
whose explicit form is reported in~\cite{Fontanelli2021} and holds true when the anchor positions are perfectly known a-priori, i.e., a map of the anchors is available.

The problem presented in this paper is different from the classic multilateration just reported, since for the problem at hand, the positions of the anchors are deployed by the robot, hence affected by uncertainty except for the very first set $\mathcal{A}_0$.
From this perspective, the problem is more similar to a Simultaneous Localisation And Mapping (SLAM) problem rather than a standard positioning problem, since the anchor map is built on the fly. Indeed, while the ranging measurements from an anchor in $\mathcal{A}_0$ is simply~\eqref{eq:rangingnoise}, from the $i$-th deployed anchor turns to
\begin{equation}
    \label{eq:rangingnoisePlaced}
    \bar{\rho}_{i,k} = \sqrt{(x_k - \hat X_i + \delta_{i_x})^2 + (y_k - \hat Y_i + \delta_{i_y})^2} + \epsilon_{i,k} ,
\end{equation}
where we denote with $\delta_i = [\delta_{i_x},\delta_{i_y}]^T$ the deployment error and with $\hat a_i = [\hat X_i, \hat Y_i]^T$ the estimated anchor position (i.e., $a_i = \hat a_i - \delta_i$). Assuming that the $i$-th anchor has been deployed at time $k T_s$, we have that $\hat a_i = \hat s_k$, hence given by~\eqref{eq:CovGDoP0}, thus affected by an uncertainty described by the covariance matrix~\eqref{eq:CovGDoP1}. In a typical SLAM problem, the first estimate of the position of a feature (which is used in the next steps as a landmark for localisation) is treated as the mean value of a random variable, usually considered as Gaussian. Applying this idea to the problem at hand, the feature estimate turns to be the anchor estimated position $\hat a_i$ and the $\delta_i$ the corresponding random variable of the uncertainty, customarily assumed with zero-mean and generated by a white stochastic process. To analyze the effect of this uncertainty, we may rewrite~\eqref{eq:rangingnoisePlaced} with its first order Taylor approximation with respect to $\epsilon_{i,k}$ and $\delta_i$, thus obtaining
\begin{equation}
\label{eq:NewRanging}
\bar{\rho}_{i,k} = \rho_{i,k} + \epsilon_{i,k} + F \delta_i = \rho_{i,k} + \eta_{i,k} ,
\end{equation}
where $F = \frac{\partial \bar{\rho}_{i,k}}{\partial \delta_i}$ is the gradient of~\eqref{eq:rangingnoisePlaced} evaluated in the mean value of $\delta_i$, i.e. $F$ is the same of $P$ in~\eqref{eq:PDoPFun}, but evaluated in $\hat a_i$. Therefore, using~\eqref{eq:NewRanging} instead of~\eqref{eq:rangingnoise}, the overall uncertainty for the ranging measurements from deployed anchors is expressed by $\eta_{i,k}$, which is a white zero-mean sequence with variance
\begin{equation}
\label{eq:deltaRV}
    \sigma_{\eta_{i,k}}^2 = \sigma_{\rho}^2 + F \Xi^{(n)} F^T ,
\end{equation}
where $\Xi^{(n)}$ is given in~\eqref{eq:CovGDoP1} (i.e., the robot position uncertainty during the placement). Since $\sigma_{\eta_{i,k}}^2 \geq \sigma_{\rho}^2$, when the $i$-th deployed anchor is used, the ranging uncertainty will be larger. For instance, assuming that at time $k T_s$ the robot uses the anchors $1$ and $2$ from $\mathcal{A}_0$ and anchors $i$ and $j$ newly placed, i.e. for which only the estimates $\hat a_i$ and $\hat a_j$ are available, we have the new form of~\eqref{eq:Nk} as % \dan{I have removed the $4$s, cause I'm not sure}.
\[
    N_k^{(n)}\!\!\! =\!\!\! 
    \begin{bmatrix}
        \sigma_{\rho}^2(\rho_{1,k}^2 \!+\! \rho_{2,k}^2) & \sigma_{\rho}^2\rho_{1,k}^2 &   \sigma_{\rho}^2\rho_{1,k}^2\\
        \sigma_{\rho}^2\rho_{1,k}^2 & \sigma_{\rho}^2\rho_{1,k}^2 \!+ \! \sigma_{\eta}^2\rho_{i,k}  & \sigma_{\rho}^2\rho_{1,k}^2\\
        \sigma_{\rho}^2\rho_{1,k}^2 & \sigma_{\rho}^2\rho_{1,k}^2  & \sigma_{\rho}^2\rho_{1,k}^2 \!+\! \sigma_{\eta}^2\rho_{j,k}
    \end{bmatrix} \!.
\]
However, the previous development along the lines of the classic SLAM approach is not entirely correct, as empirically proved in Section~\ref{sec:SimulationResults}. Indeed, $\delta_i$ should not be considered as a white random variable with zero-mean and covariance~\eqref{eq:CovGDoP1} but, instead, as a realization of a random variable at time $k T_s$, i.e., a realization of the random variable modelling the robot positioning uncertainty, hence an unknown but constant offset.
With this assumption, a typical non Bayesian approach as the nonlinear WLS can be adopted. More precisely, given at least three consecutive ranging measurements $\bar{\rho}_{i,k}$, $\bar{\rho}_{i,k+1}$ and $\bar{\rho}_{i,k+2}$ described in~\eqref{eq:rangingnoisePlaced}, the value of $\delta_i$ is given by
\begin{equation}
\label{eq:CovGDoP7}
        \hat\delta_i \!=\!\argmin\limits_{(\delta_{i_x}, \delta_{i_y})} \sum_{j=k}^{k+2}[(\hat x_j - \hat X_i + \delta_{i_x})^2 \!+ (\hat y_j - \hat Y_i + \delta_{i_y})^2 \!- \bar{\rho}_{i,j}^2]^2 .
\end{equation}
This way, the offset $\delta_i$ induced in the anchor placement by the robot position uncertainty $\tilde s_k$ can be estimated and, hence, removed from the anchor estimated position $\hat a_i$ by means of this nonlinear unconstrained regression problem, as shown in the next section. It is important to remark that the PDoP in~\eqref{eq:PDoPFun} does not consider the effect of the offset on the deployed anchors position. These effects are voluntarily neglected because their contributions to the estimation of the PDoP generate. In the worst case of an offset in the order of tens of centimetres, a difference with the actual PDoP is less than $3\%$ of $p^m$. Therefore, being the offset errors after~\eqref{eq:CovGDoP7} of the order of few centimetres, we simply impose a PDoP threshold of $95\%$ of $p^m$ to account for those effects and design a conservative approach.

\section{Simulations and Experiments}
\label{sec:SimulationResults}

\begin{figure}[t]
    \centering
    \includegraphics[width=0.8\columnwidth]{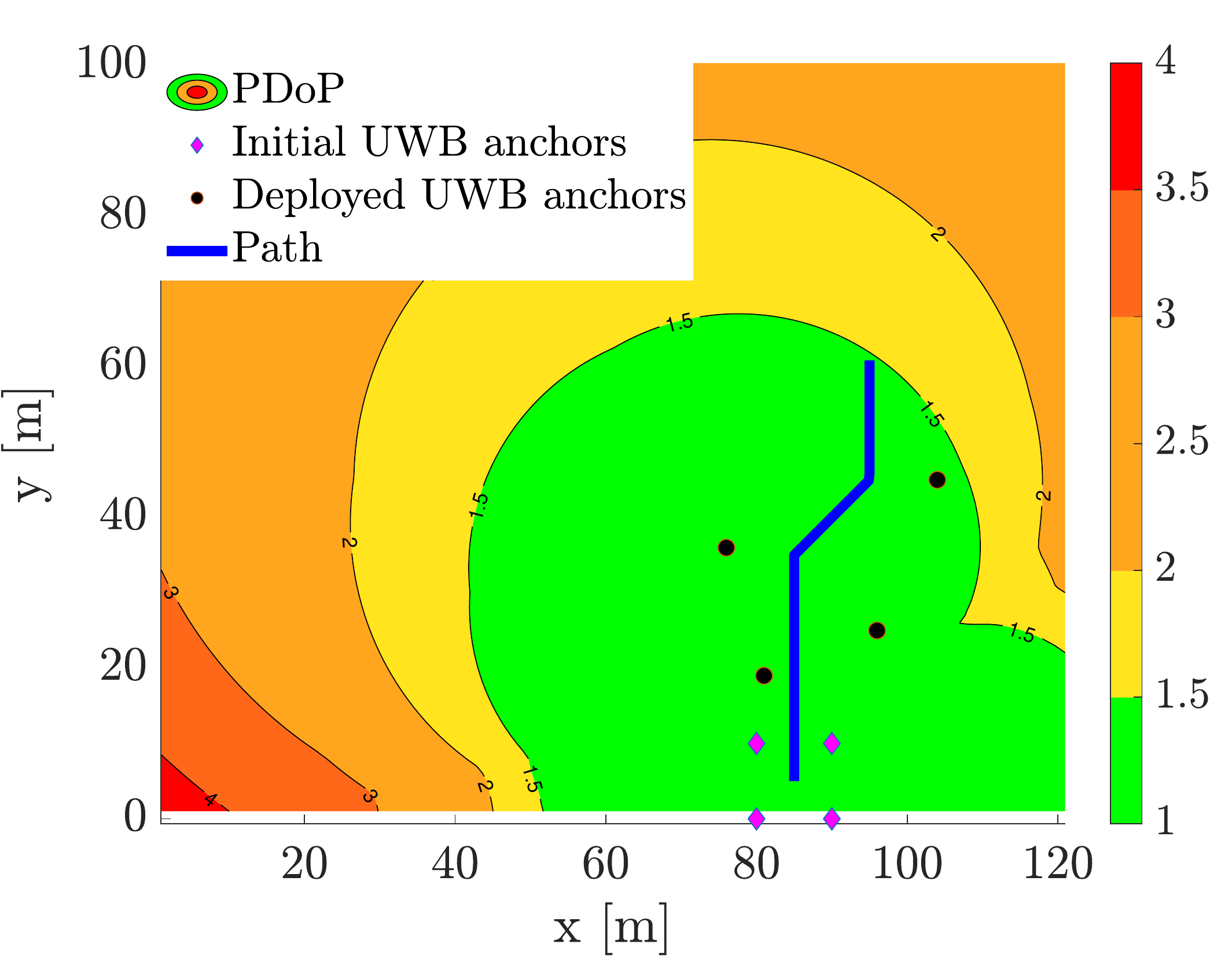}
    \caption{PDoP level curves computed using $\mathcal{D}_k(s_k)$, where $s_k$ covers the entire map.}
    \label{fig:HeatmapGANP}
\end{figure}

\begin{figure}[t]
    \centering
    \includegraphics[width=0.8\columnwidth]{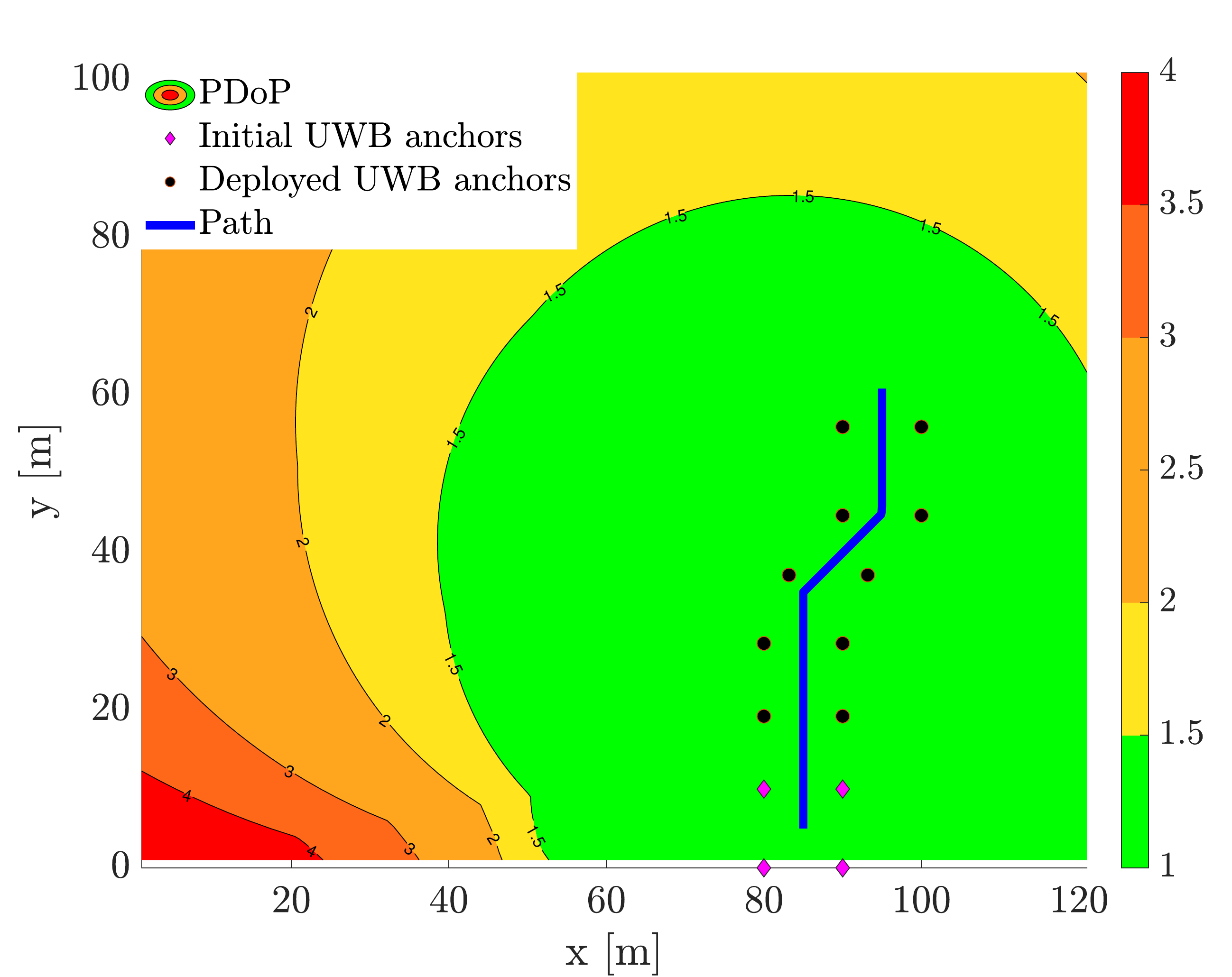}
    \caption{PDoP level curves computed using $\mathcal{D}_k(s_k)$ for the straightforward algorithm sketched at the beginning of Section~\ref{sec:AnchorDeployment}.}
    \label{fig:HeatmapTrivial}
\end{figure}

\begin{figure}[t]
    \centering
    \includegraphics[width=1\columnwidth]{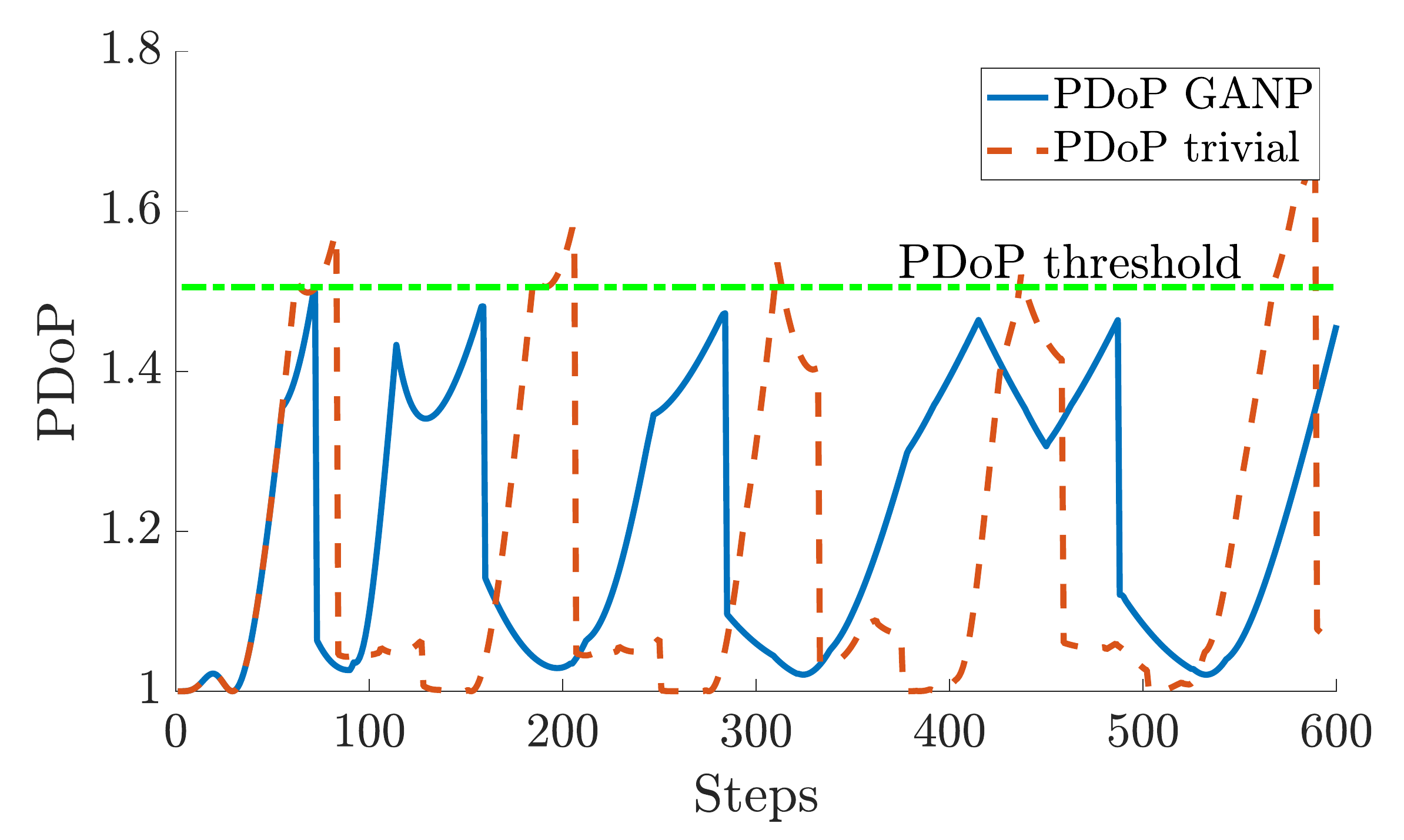}
    \caption{PDoP evolution along the simulations in Figure~\ref{fig:HeatmapGANP} (GANP) and Figure~\ref{fig:HeatmapTrivial} (trivial), respectively.}
    \label{fig:PDoPComparison}
\end{figure}

\begin{figure}[t]
    \centering
    \includegraphics[width=1\columnwidth]{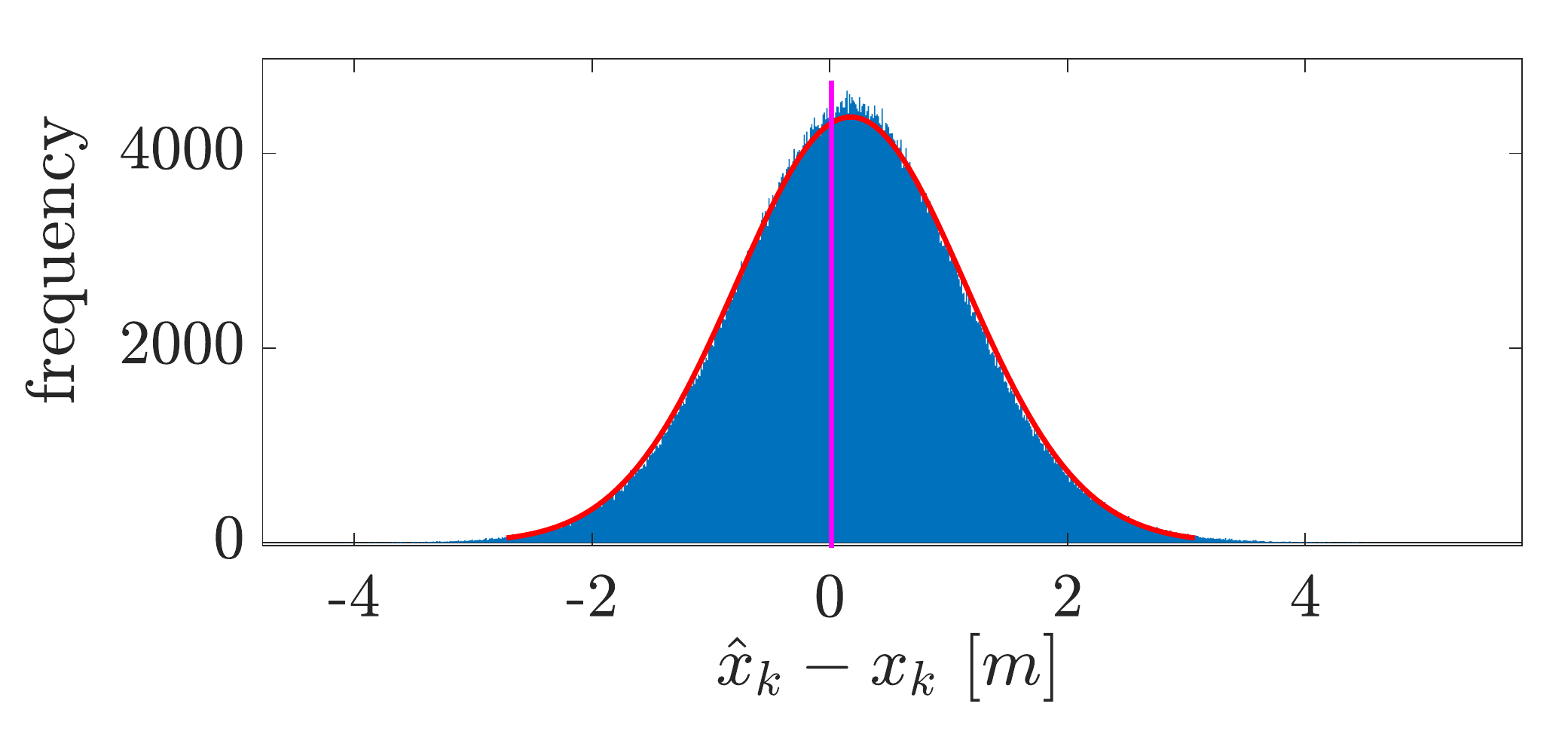}
    \caption{Monte Carlo trials for the placement problem. When the placement error $\delta_i$ is treated as a random variable, a bias of about $15$~cm is induced (dashed line), while if it is treated as an unknown but constant quantity estimated through~\eqref{eq:CovGDoP7} (solid line), the estimator is practically unbiased (bias around $1$~mm).}
    \label{fig:MonteBias}
\end{figure}

To evaluate the effectiveness of the GANP algorithm, we first present here the simulation results. We assume that the maximum ranging distance is ${\rho}^m = 60$~m, which is derived by the hardware specification of the Decawave DWM1001 UWB anchors. To fine-tune the parameters of the GANP algorithm, i.e., the area width $w$, the number of subareas $n$ and the horizon of the prediction $r$, we report here an analysis based on the Taguchi Orthogonal Array (OA) design~\cite{taguchi1986introduction}. To this end, we impose the maximum PDoP value to be ${p}^m = 1.5$ (a value guaranteeing low positioning uncertainty and a sufficiently large feasible placement region) and an exploration path length of approximately $60$~m. The result of the analysis, reported in terms of the performance indices number of anchors $m$, travelled distance $d_t$ and computational time $c_t$, is subsumed in Table~\ref{tab:DoE}.
%-%
\begin{table}[t]
    \caption{Performance of the GANP algorithm versus parameter choices.}
    \label{tab:DoE}
    \centering
    \begin{tabular}{|ccc|ccc|}
        \hline
        \multicolumn{3}{|c|}{Parameters} & \multicolumn{3}{c|}{Performance indices} \\
        \hline
        $w$ [m] &  $r$ [m] &  $n$ & $m$ & $d_t$ [m] & $c_t$ [s] \\
        \hline 
        10 & 10 & 2 & 12 & 138 & 1118\\
        10 & 20 & 4 & 12 & 140 & 1007\\
        10 & 30 & 3 & 11 & 145 & 450\\
        30 & 10 & 4 & 10 & 166 & 307\\
        30 & 20 & 3 & 9 & 154 & 362\\
        30 & 30 & 2 & 9 & 146 & 895\\
        50 & 10 & 3 & 9 & 170 & 680\\
        50 & 20 & 2 & 9 & 180 & 650\\
        50 & 30 & 4 & 9 & 176 & 526\\
        \hline
    \end{tabular}
\end{table}
%-%
It is evident that a larger area minimises the number of deployed anchors, since the feasible deploying space increases, at the price of a higher travelled distance. Instead, while the computation time clearly increases for larger areas (i.e., a larger space to explore for the GA algorithm), too small areas may imply difficulties in the search for a suitable solution. Hence the computation time increases as well. An optimal choice of the parameters would lead to the optimization of all the performance indices at once, which is hardly possible for the contrasting goals explained. Therefore, we compute the choice of the parameters $w$, $r$ and $n$ minimising each selected performance index at once and then compute the average among them, as reported in Table~\ref{tab:FactorComb}.
%-%
\begin{table}[t]
    \caption{Optimal choices of the parameters.}
    \label{tab:FactorComb}
    \centering
    \begin{tabular}{|c|ccc|}
        \hline
        Performance indices & $w$ [m] & $r$ [m] & $n$ \\
        \hline 
        $m$ & 50 & 30 & 3\\
        $d_t$ & 10 & 30 & 2\\
        $c_t$ & 30 & 30 & 3\\
        \hline
        \textbf{Average value} & 27 & 30 & 3\\
        \hline
    \end{tabular}
\end{table}
%-%
The placement obtained with the depicted tuning is then applied to a simulation example, thus obtaining the level curves of the PDoP reported in Figure~\ref{fig:HeatmapGANP}.
%-%
%-%
In this case, just $4$ new anchors have been added to cover the entire exploration path and respecting the PDoP limit ${p}^m = 1.5$. For comparison, Figure~\ref{fig:HeatmapTrivial} reports the same scenario assuming the trivial approach sketched at the beginning of Section~\ref{sec:AnchorDeployment}. 
%-%
%-%
It is evident that, albeit simple, this algorithm implies a waste of resources, imposing the PDoP region $g(\mathcal{D}_k(s_k), s_k) \geq p^m$ to be too wide comapred to the exploration task. Moreover, as it can be observed from Figure~\ref{fig:PDoPComparison}, the PDoP constraint is not always verified along the exploration path or the placement path for the simple approach, while it is strictly satisfied for the GANP algorithm.
%-%
%-%
As a final simulation test, we verified that the SLAM-like assumption of the anchor estimated positions $\hat a_i$ cannot be considered as a random variable, as discussed in Section~\ref{subsec:AnchorDeployment}. To empirically prove this fact, we have carried out $10^6$ Monte Carlo trials where $\delta_i$ uncertainty is treated as a random variable contributing to the random, zero-mean white noise in~\eqref{eq:deltaRV} and hence applying the multilateration~\eqref{eq:CovGDoP0}, which results in the position uncertainty in Figure~\ref{fig:MonteBias}, dashed line.  
%-%
%-%
As can be noticed, this assumption end up with a non-negligible bias on the estimates of the estimated position $\hat s_k$ (the Figure~\ref{fig:MonteBias} reports the bias on the $\hat x_k$ axis, but it acts similarly on $\hat y_k$). Consequently, the bias should be treated as a constant but unknown quantity using~\eqref{eq:CovGDoP7}, thus resulting in the unbiased estimation uncertainty of Figure~\ref{fig:MonteBias}, solid line. In the next section, this phenomenon is additionally highlighted in the experiments.

\subsection{Experimental results}

\begin{figure}[t]
  \begin{tabular}{cc}
    \centering
    \includegraphics[width=0.45\columnwidth]{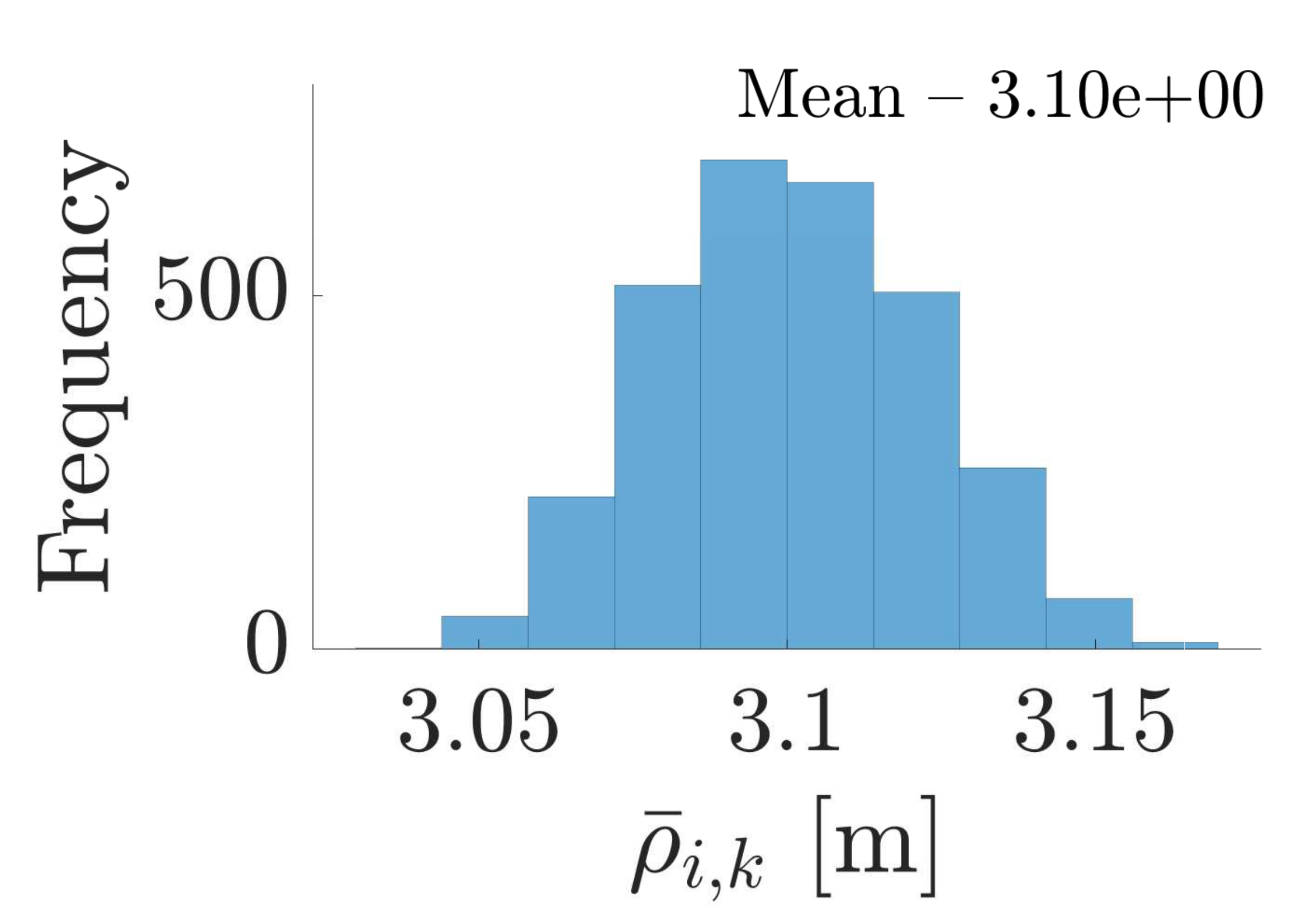} &
    \includegraphics[width=0.45\columnwidth]{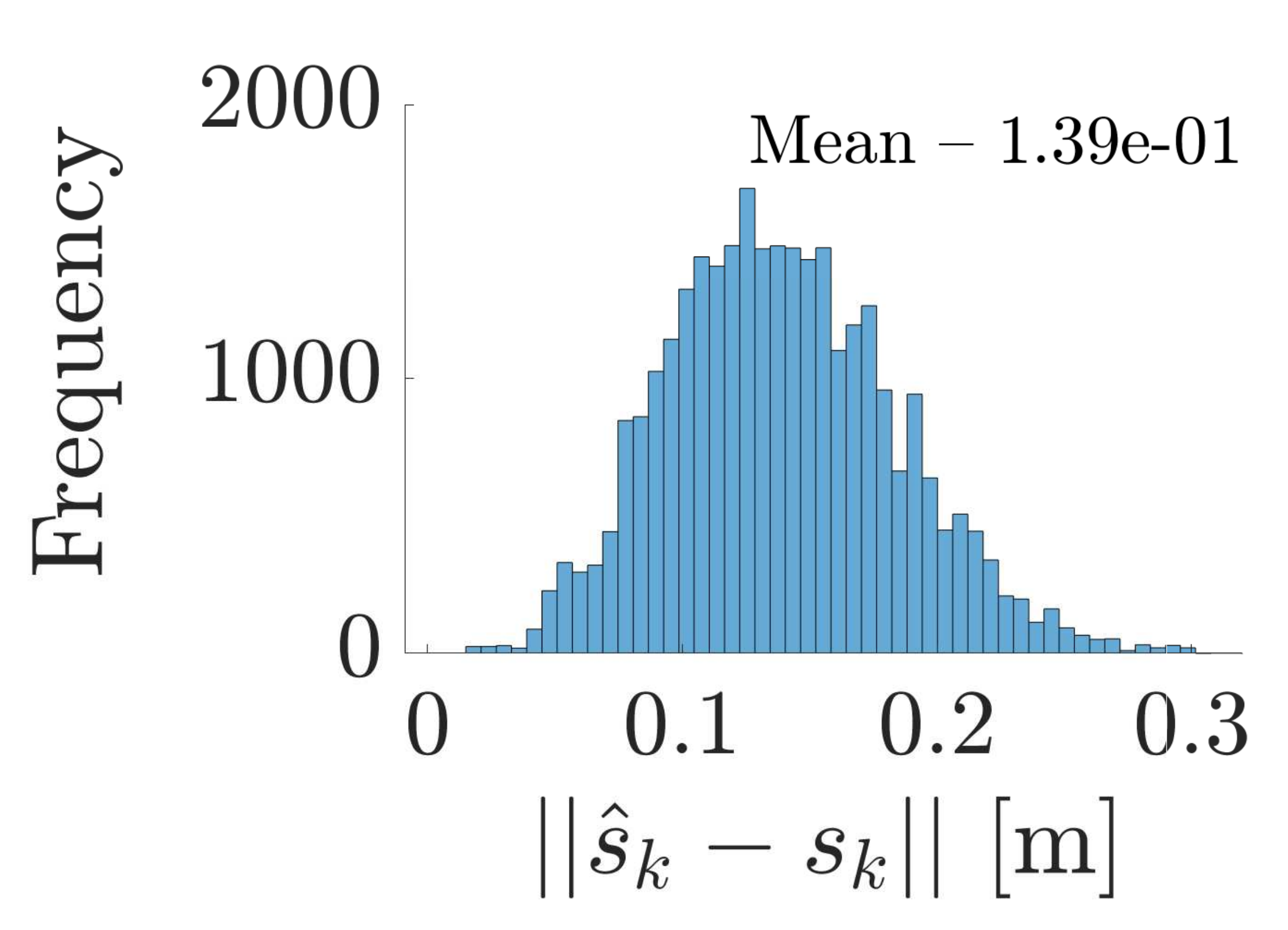}
    \\
    (a) & (b)
  \end{tabular}
  \caption{(a) Characterisation of the ranging measurements $\bar\rho_{i,k}$ by means of an histogram obtained with 3000 consecutive measurements (b) Histogram of positioning error with 30000 consecutive measurements}
  \label{fig:histogram3meters}
\end{figure}

\begin{figure}[t]
    \centering
    \includegraphics[width=\columnwidth]{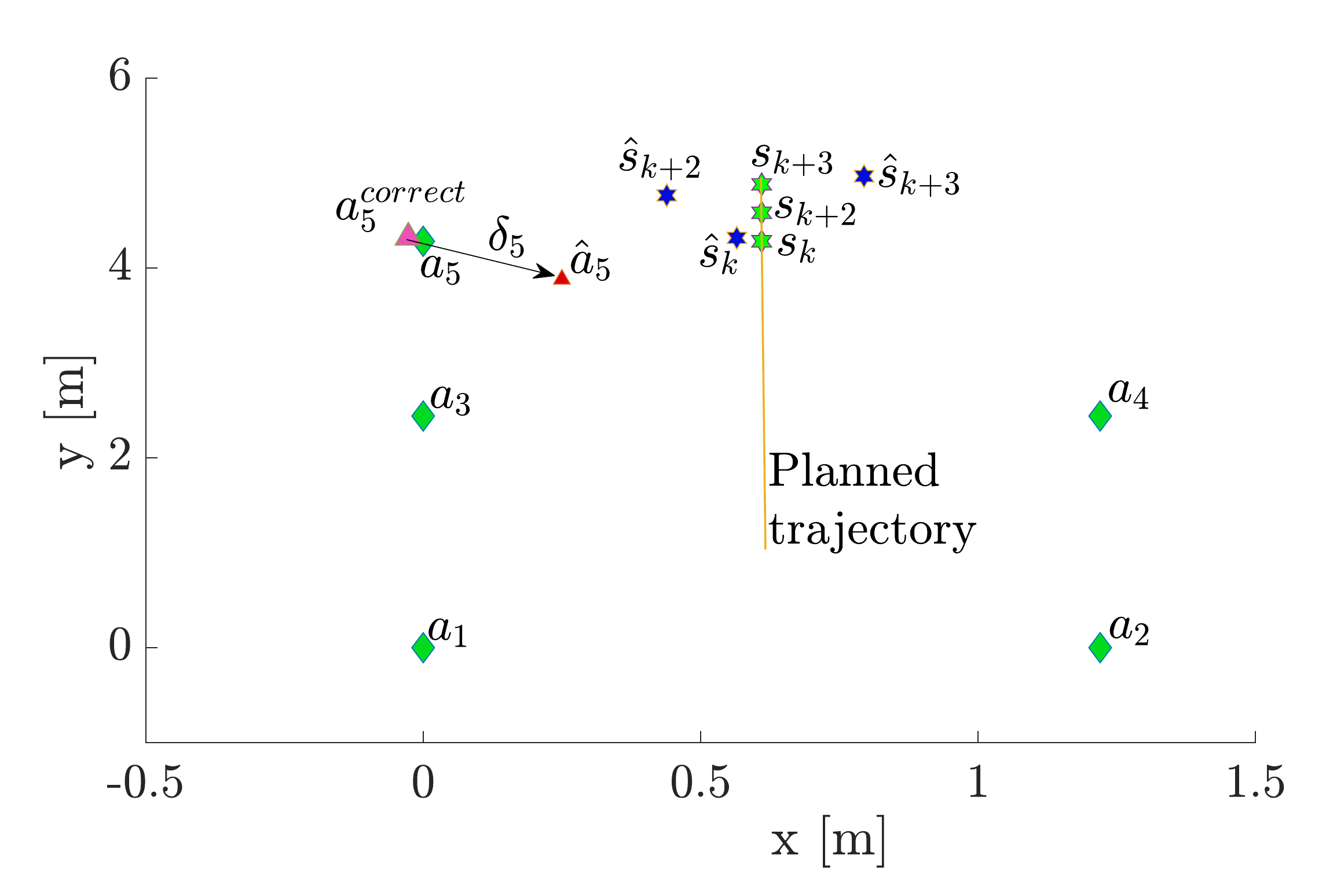}
    \caption{Experimental results for the bias $\delta_5$ compensation.}
    \label{fig:ExperimentLM} 
    %\vspace{-5mm}
\end{figure}

\begin{figure}[!t]
    \centering
    \includegraphics[width=0.85\columnwidth]{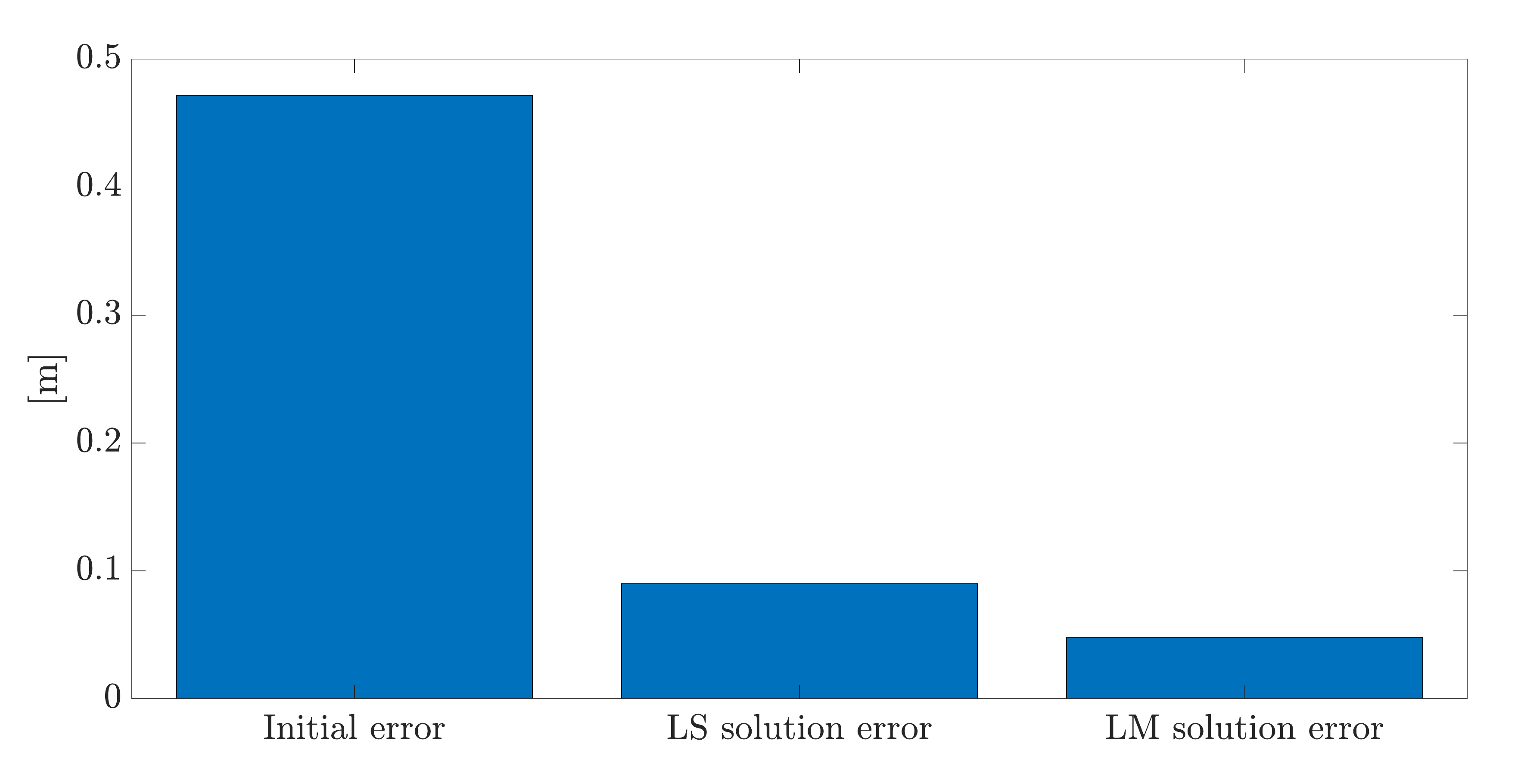}
    \caption{Comparison between the LM and the LS solution to~\eqref{eq:CovGDoP7} with respect to the initial deployment error.}
    \label{fig:CompLM_LS}
\end{figure}

To test the algorithm on an actual set-up, we first characterise the UWB anchors at disposal. To this end, we carried out a Type A analysis~\cite{kirkup2006introduction}, collecting at first repeated ranging measurements from known distances, i.e., at $1$, $3$ and $7$ meters. As an example, the histogram of the ranging measurements $\bar\rho_{i,k}$ in~\eqref{eq:rangingnoise} collected from an UWB anchor Decawave DWM1001 at a distance of $3$~m is reported for reference in Figure~\ref{fig:histogram3meters}-a. Figure~\ref{fig:histogram3meters}-b reports the histogram of the error on the position.
%-%
%-%
Albeit all the available anchors behave similarly and with a relatively small variance $\sigma_\rho^2$, they all exhibit an approximately linear dependency on the actual distance $\rho_{i,k}$, as reported in Table~\ref{tab:StatProp} for the three sampled distances.
%-%
\begin{table}[t]
    \caption{Characterisation of bias and standard deviation of the ranging measurements}
    \label{tab:StatProp}
    \centering
    \begin{tabular}{|c|cc|}
        \hline
        $\rho_{i,k}$ [m] & Bias [m] & $\sigma_\rho$ [m] \\
        \hline 
        1 & 0.06 & 0.029 \\
        3 & 0.10 & 0.0231 \\
        7 & 0.18 & 0.16 \\
        \hline
    \end{tabular}
\end{table}
%-%
We may noticed a slight increase of the bias and of the standard deviation $\sigma_\rho$, which can be compensated with a simple linear fitting model.
Since we do not have a large arena to test the system, we test the GANP placement algorithm forcing the anchors to be closed to each other by selecting ${p}^m = 2$ to be above the minimum PDoP value obtained for the known first four anchors, which was $g(\mathcal{D}_k(s_0), s_0) = 1.1$ ($\mathcal{D}_k(s_0) = \{a_1, \dots, a_4\}$ in Figure~\ref{fig:ExperimentLM}). 
%-%
%-%
As stated previously, once the new anchor has been deployed in position $a_5$ and due to the positioning uncertainty of the UAV, the robot actually believes that the anchor is in $\hat a_5$ (Figure~\ref{fig:ExperimentLM}). After the placement, the UAV comes back to the exploration path (solid line in Figure~\ref{fig:ExperimentLM}) and, as described in Section~\ref{subsec:AnchorDeployment}, it stores three consecutive estimated positions, namely $\hat s_k$, $\hat s_{k+1}$ and $\hat s_{k+2}$, which actually corresponds to the ideal values to be reached $s_k$, $s_{k+1}$ and $s_{k+2}$ (see Figure~\ref{fig:ExperimentLM} for reference). To estimate the bias $\delta_5$, the robot collects $10$ consecutive ranging measurements $\overline{\rho}_{5,k}$ from position $s_k$ and then compute the average. The process is then repeated from $s_{k+1}$ and $s_{k+2}$. The solution to~\eqref{eq:CovGDoP7} is then obtained using the Levenberg-Marquardt (LM) algorithm applied to the averages, thus obtaining the corrected location $a_5^{correct}$ of Figure~\ref{fig:ExperimentLM}, exhibiting a far reduced bias compared to $\hat a_5$.  The method thus described has been compared with a linearized least square (LS) solution to~\eqref{eq:CovGDoP7} on such experimental data, which results in the comparison of Figure~\ref{fig:CompLM_LS}.
%-%
%-%
Of course the iterative and incrementally precise approach of LM gives better results than LS for the bias estimation. Similarly, LM performs better of LS also for the standard positioning problem using multilateration. However, since the algorithm can be executed on board the vehicle and with constrained resources, the LM should be adopted with parsimony (its computation times is about $40$ times compared to a linearized LS). As a consequence, we decided to keep the LM solution uniquely for the bias estimation problem. 

\section{Conclusion}
\label{sec:Conclusion}

This paper presented the development of a self-deployable UWB infrastructure by robots, with simulations and experiments done on small indoor UAVs to confirm the effectiveness of our approach. Starting from a minimal pre-deployed infrastructure,  the robot can extend the UWB anchors geometry while exploring the environment during the mission.  Clearly,  uncertainty could increment while expanding the UWB infrastructure. To address this concern, we developed a genetic algorithm to compute the optimal placement of new anchors using the  Geometric Dilution of Precision (GDoP). Simulation and experimental results demonstrated that the positioning algorithm uncertainty is always kept under the threshold required by the user. Future research threads will be devoted to apply the solution to a team of heterogeneous robots and to extend the analysis to the localisation problem, where the motion of the robot and its dynamic should be considered in the placement problem.

%\newpage

\bibliographystyle{IEEEtran}
%\bibliography{biblio}

\vspace{-28em}

\begin{IEEEbiography}[{\includegraphics[width=1in,height=1.25in,clip,keepaspectratio]{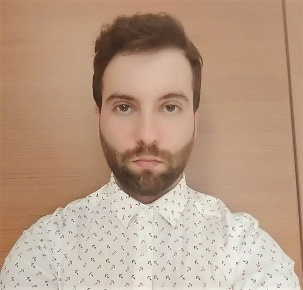}}]{Luca Santoro}  is currently Ph.D. Student at the Department of Industrial Engineering, University of Trento, Italy. He obtained his M.S in the field of Electronics and Robotics Engineering in 2020. His research interests encompass the development of autonomous UAVs, the investigation machine learning functionalities applied to resource constrained embedded platforms, distributed and real-time estimation and control, localisation algorithms and clock synchronisation algorithms.
\end{IEEEbiography}

\vspace{-28em}
%\newpage

%
\begin{IEEEbiography}[{\includegraphics[width=1in,height=1.25in,clip,keepaspectratio]{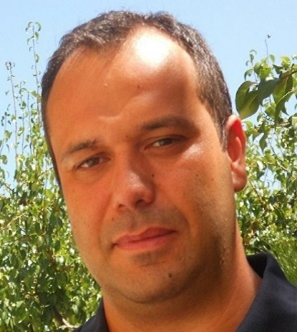}}]{Davide Brunelli} (Senior Member,~IEEE) received the M.S. (cum laude) and Ph.D. degrees in electrical engineering from the University of Bologna, Bologna, Italy, in 2002 and 2007, respectively. He is currently an associate professor at the University of Trento, Italy.  His research interests include IoT and distributed lightweight unmanned aerial vehicles, the development of new techniques of energy scavenging for low-power embedded systems and energy-neutral wearable devices, Drones, UAVs and Machine Learning.  He was leading industrial cooperation activities with Telecom Italia, ENI, and STMicroelectronics. He has published more than 200 papers in international journals and proceedings of international conferences. He is an ACM member and an IEEE Senior member.
\end{IEEEbiography}
%

%\vspace{-25em}
\newpage

\begin{IEEEbiography}[{\includegraphics[width=1in,height=1.25in,clip,keepaspectratio]{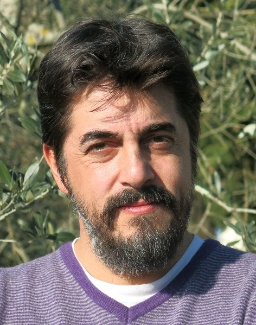}}]{Daniele Fontanelli} (Senior Member,~IEEE) received the M.S. degree in Information Engineering in 2001, and the Ph.D. degree in Automation, Robotics and Bioengineering in 2006, both from the University of Pisa, Pisa, Italy. He was a Visiting Scientist with the Vision Lab of the University of California at Los Angeles, Los Angeles, US, from 2006 to 2007. From 2007 to 2008, he has been an Associate Researcher with the Interdepartmental Research Center “E. Piaggio”, University of Pisa. From 2008 to 2013 he joined as an Associate Researcher the Department of Information Engineering and Computer Science and from 2014 the Department of Industrial Engineering, both at the University of Trento, Trento, Italy, where he is now an Associate Professor and he is leading the EITDigital international Master on “Autonomous Systems”. He has authored and co-authored more than 150 scientific papers in peer-reviewed top journals and conference proceedings. He is currently an Associate Editor in Chief for the IEEE Transactions on Instrumentation and Measurement and an Associate Editor for the IEEE Robotics and Automation Letters and for the IET Science, Measurement \& Technology Journal. He has also served in the program committee of different conferences in the area of measurements and robotics. His research interests include distributed and real-time estimation and control, human motion models, localisation algorithms, synchrophasor estimation, clock synchronisation algorithms, resource aware control, wheeled mobile robots control and service robotics.
\end{IEEEbiography}

\end{document}